\documentclass[runningheads]{llncs}
\usepackage[T1]{fontenc}

\usepackage{float}
\usepackage{amsmath}
\usepackage{makecell}
\usepackage{multirow}
\usepackage{orcidlink}

\usepackage{amssymb}
\usepackage{orcidlink}

\usepackage{graphicx,verbatim}
\usepackage{graphicx}
\usepackage{float}
\usepackage{booktabs}
\usepackage{bm}
\usepackage{hyperref}

\usepackage{array}

\usepackage{marvosym}
\begin{document}
\title{CoDiR: Confidence-Guided Diffusion Refinement for Semi-Supervised Histopathology Segmentation}
\titlerunning{CoDiR: Confidence-Guided Diffusion Refinement}

%
\makeatletter
\newcommand{\printfnsymbol}[1]{%
  \textsuperscript{\@fnsymbol{#1}}%
}
\makeatother
\author{
Hoai Nhan Pham\inst{1}\thanks{Equal contribution} \and
Dang-Nguyen Bui\inst{2}\printfnsymbol{1} \and
Le-Van Thai\inst{1}\printfnsymbol{1} \and
Thanh-Hiep Vo\inst{3} \and
Lan Anh Dinh Thi\inst{4} \and
Tien Dat Nguyen\inst{1} \and
Duy-Dong Nguyen\inst{1} \and
Ngoc Lam Quang Bui\inst{5} \and
Tam Tran\inst{6}\textsuperscript{(\Letter)}\orcidlink{0009-0000-9702-425X} \and
Zhi Huang\inst{7}\textsuperscript{(\Letter)}\orcidlink{0000-0001-6982-8285}
}

\authorrunning{Hoai Nhan Pham et al.}

\institute{
AI VIETNAM Lab, Vietnam\\
\and
Portland Community College, USA\\
\and
University of Science, VNU-HCM, Ho Chi Minh City, Vietnam\\
\and
Hanoi University of Science and Technology, Vietnam\\
\and
Department of Mechanical System Engineering, Jeonbuk National University, Republic of Korea\\
\and
John T. Milliken Department of Medicine, Washington University School of Medicine, Saint Louis, MO, USA\\
\email{tamt@wustl.edu}
\and
Perelman School of Medicine, University of Pennsylvania, USA\\
\email{zhi.huang@pennmedicine.upenn.edu}
}

\maketitle

\begin{abstract}
Semi-supervised histopathology segmentation is challenging due to scarce annotations and unreliable pseudo-labels in ambiguous gland regions. To address this problem, we propose Confidence-Guided Diffusion Refinement (CoDiR), a semi-supervised framework that combines a Mean Teacher segmentation model with diffusion-based pseudo-label refinement. Given an unlabeled image, the teacher first produces a soft prediction, and only low-confidence regions are refined by a conditional diffusion model trained to capture plausible mask structures from labeled data. The refined mask is then fused with reliable teacher predictions and used to train the student with confidence weighting and consistency regularization. On the GlaS and CRAG datasets CoDiR reaches 88.09\% and 89.83\% mDice with 10\% labeled data, and 89.19\% and 90.29\% mDice with 20\%, matching or exceeding the strongest published method on seven of the eight benchmark metrics. Ablations attribute the largest single contribution to the refinement module, which adds +6.36\% mDice over the Mean Teacher baseline. The implementation code is publicly available at: \url{https://github.com/vongla345/codir}
\end{abstract}
\section{Introduction}
Medical image segmentation plays a crucial role in computer-aided diagnosis and pathological analysis by enabling accurate delineation of anatomical structures \cite{long2015fully,ronneberger2015u}. However, acquiring pixel-level annotations for histopathology images is costly and labor-intensive, requiring expert pathologists to label complex tissue structures with high morphological variability and staining-induced appearance changes \cite{shen2023co,kapse2024si,wu2025learning}.

Semi-supervised learning (SSL) alleviates annotation costs by leveraging abundant unlabeled data alongside limited labeled samples \cite{han2024deep}. Teacher--student frameworks based on pseudo-labeling and consistency regularization have achieved strong performance in medical image segmentation \cite{sohn2020fixmatch,yu2019uncertainty,chen2021semi,sun2024corrmatch}. However, inaccurate pseudo-labels may introduce structural and boundary errors that are repeatedly reinforced during training, leading to confirmation bias and degraded segmentation quality \cite{arazo2020pseudo,liu2024diffrect}.

Meanwhile, pathology foundation models such as UNI \cite{chen2024uni} have demonstrated strong transferability across downstream pathology tasks. Building on these pretrained representations, our earlier work, UniSemAlign~\cite{van2026unisemalign}, introduced prototype- and text-guided semantic alignment for semi-supervised histopathology segmentation. While it improves representation quality and feature discrimination, it constrains pseudo-labels only indirectly through feature alignment and does not explicitly refine erroneous mask regions.

To address these limitations, we propose Confidence-Guided Diffusion Refinement (CoDiR), a semi-supervised histopathology segmentation framework that leverages pathology foundation representations and diffusion-guided pseudo-label refinement. By selectively correcting unreliable pseudo-label regions, CoDiR provides more reliable supervision for unlabeled data. Extensive experiments on the CRAG and GlaS benchmarks demonstrate the effectiveness of the proposed approach.

Our main contributions are as follows:

\begin{itemize}
\item We propose \textbf{CoDiR}, a semi-supervised histopathology segmentation framework that combines a frozen UNI encoder, a DeepLabV3-style decoder \cite{chen2017rethinking}, and Mean Teacher learning \cite{tarvainen2017mean}. Where our earlier UniSemAlign framework~\cite{van2026unisemalign} improves pseudo-label quality indirectly through semantic alignment, CoDiR acts on the mask itself, correcting unreliable regions with a learned structural prior.
\item We introduce a \textbf{confidence-gated diffusion refinement} module that selectively corrects low-confidence regions while preserving reliable teacher predictions. Compared with DiffRect \cite{liu2024diffrect} and SDEdit-style editing \cite{meng2021sdedit}, our approach refines only uncertain pixels with a conditional diffusion prior, reducing confirmation bias and improving boundary quality.
\item We evaluate CoDiR on the GlaS and CRAG benchmarks, where it achieves consistent improvements over recent semi-supervised methods.
\end{itemize}


\section{Methodology}

\begin{figure}[t!]
    \includegraphics[width=\textwidth]{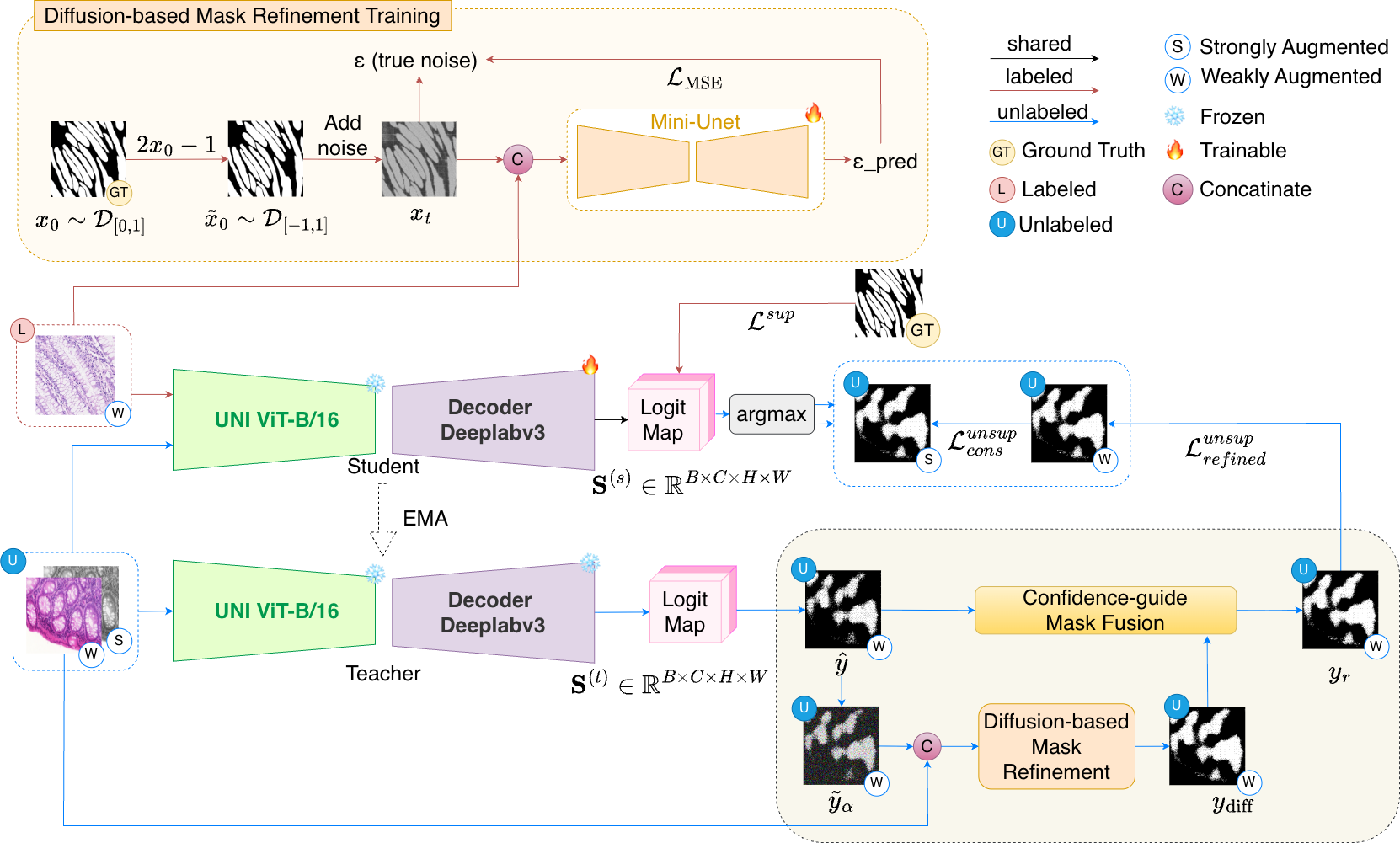}
    \caption{Overview of CoDiR. The student segmentation network is trained with labeled supervision and pseudo-labels generated by an EMA teacher. Meanwhile, the diffusion model is trained on labeled data and used within the same training loop to refine low-confidence regions of teacher pseudo-masks. The refined outputs are then fused with high-confidence teacher predictions to produce the final pseudo-labels for unlabeled learning.} \label{fig_overview}
\end{figure}

\paragraph{Overview.} We propose a semi-supervised histopathology segmentation framework that integrates a Mean Teacher segmentation model with a diffusion-based pseudo-label refinement module. As illustrated in Fig.~\ref{fig_overview}, the framework consists of a segmentation network trained under the Mean Teacher paradigm~\cite{tarvainen2017mean} and a diffusion model~\cite{ho2020denoising}. The segmentation model is supervised using labeled data and learns from unlabeled data via teacher-generated pseudo-labels. Meanwhile, the diffusion model is trained on labeled data to model the conditional distribution \(p(y|I)\) of segmentation masks. During training, the teacher produces soft pseudo-masks for unlabeled images, and only low-confidence regions are refined by the diffusion model before being fused with high-confidence predictions to form the final pseudo-label \(y_r\).

\subsection{Segmentation Model}

\noindent\textbf{Visual Encoder.}
Given an input image
\(I \in \mathbb{R}^{B \times 3 \times H \times W}\),
a pretrained UNI ViT-L/16 \cite{chen2024uni} encoder extracts patch-level visual features:

\begin{equation}
    \mathbf{F}_{img} = \text{UNI}(I) \in \mathbb{R}^{B \times N \times d_{img}}
\end{equation}

where \(N=\frac{H}{16} \times \frac{W}{16}\) and \(d_{img}=1024\). The CLS token is discarded, and the remaining patch tokens are reshaped into a 2D feature grid.

\noindent\textbf{Decoder.}
The feature \(\mathbf{F}_{img}\) passed into a DeepLabV3-style decoder~\cite{chen2017rethinking}, which aggregates multi-scale context using ASPP and upsamples the features to produce a per-pixel segmentation logit map \(\mathbf{S} \in \mathbb{R}^{B \times 1 \times H \times W} \). 

\noindent\textbf{Mean Teacher.}
The framework adopts a Teacher-Student paradigm~\cite{tarvainen2017mean}. The Student \(f_{\theta}\) is updated via gradient descent, while the Teacher \(f_{\bar{\theta}}\) maintains an exponential moving average (EMA) of the Student weights:

\begin{equation}
    \bar{\theta} \xleftarrow{} \alpha\bar{\theta} + (1-\alpha)\theta
\end{equation}

where \(\alpha\) is the EMA decay coefficient. The Teacher parameters are not directly optimized and serve solely as a pseudo-label generator.

\subsection{Diffusion-based Refinement}

Teacher predictions provide useful supervision, but they are often unreliable in ambiguous regions and may propagate confirmation bias when used directly as pseudo-labels~\cite{liu2024diffrect,arazo2020pseudo}. To address this issue, we introduce a diffusion-based refinement module that selectively corrects low-confidence regions while preserving reliable predictions.

\noindent\textbf{Diffusion model training.}
The diffusion model $\epsilon_{\phi}$ is trained exclusively on the labeled subset $\mathcal{D}_l$ to model the conditional mask distribution $p(y \mid I)$. Following the DDPM formulation~\cite{ho2020denoising}, it learns to denoise corrupted segmentation masks conditioned on the input image, thereby capturing a structural prior over plausible object shapes. To mitigate overfitting under extreme label scarcity, we use a low-capacity mini-UNet and diffusion-induced noise regularization; refinement is further confined to low-confidence regions to avoid corrupting reliable structures and suppress hallucinations.

\noindent\textbf{Confidence-guided refinement.}
For an unlabeled image \(I_u\), the teacher predicts a soft pseudo-mask
\begin{equation}
\hat{y} = \sigma\!\left(f_{\bar{\theta}}(\mathcal{A}_w(I_u))\right) \in [0,1],
\end{equation}
with confidence
\begin{equation}
c_{\text{teacher}} = \max(\hat{y}, 1-\hat{y}),
\end{equation}
where low-confidence regions are identified by an adaptive threshold \(\tau\) following prior confidence-based pseudo-labeling strategies~\cite{zhang2023ctanet}.

To refine these regions, we adopt a partial noising strategy inspired by SDEdit~\cite{meng2021sdedit}. The pseudo-mask is perturbed to an intermediate noise level \(\alpha_{\mathrm{n}}\) to obtain \(\tilde{y}_{\alpha_{\mathrm{n}}}\), then denoised for \(K\) DDPM reverse steps conditioned on \(I_u\):
\begin{equation}
y_{\text{diff}} = \text{Denoise}_K(\tilde{y}_{\alpha_{\mathrm{n}}}, I_u; \epsilon).
\end{equation}

The confidence map is thresholded to produce a binary refinement mask \(m\), and the refined pseudo-mask is computed as
\begin{equation}
y_r = m \odot y_{\text{diff}} + (1-m)\odot \hat{y},
\end{equation}
where \(m \in \{0,1\}^{H \times W}\) denotes pixels with \(c_{\text{teacher}} < \tau\).

\noindent\textbf{Confidence weighting.}
To suppress noisy supervision, each pixel is assigned a confidence weight based on the agreement between the teacher and diffusion outputs~\cite{li2023confidence}:
\begin{equation}
w = c_{\text{teacher}} \cdot c_{\text{diffusion}}^{\text{eff}},
\end{equation}
with
\begin{equation}
c_{\text{diffusion}}^{\text{eff}} = m \odot \max(y_{\text{diff}}, 1-y_{\text{diff}}) + (1-m)\odot c_{\text{teacher}}.
\end{equation}
Thus, diffusion confidence is used in refined regions, while teacher confidence is retained elsewhere, so that reliable pixels contribute more strongly to the unsupervised loss.

\subsection{Training Objectives}
The overall objective is defined as
\begin{equation}
\mathcal{L}
=
\frac{1}{2}
\left(
\mathcal{L}_{\mathrm{sup}}
+
\mathcal{L}_{\mathrm{unsup}}
\right)
\end{equation}
The supervised loss is computed on labeled data, while the unsupervised loss is applied to unlabeled samples using diffusion-refined pseudo-labels.

\paragraph{Supervised loss.}
For labeled images, the binary student prediction \(S\) is supervised with the ground-truth mask using a combination of pixel-wise and region-level losses:
\begin{equation}
\mathcal{L}_{\mathrm{sup}}
=
\lambda_1 \mathcal{L}_{\mathrm{CE}}
+
\lambda_2 \mathcal{L}_{\mathrm{Dice}}
+
\lambda_3 \mathcal{L}_{\mathrm{clDice}}
+
\lambda_4 \mathcal{L}_{\mathrm{Boundary}}
\end{equation}
where \(\mathcal{L}_{\mathrm{clDice}}\)~\cite{shit2021cldice} preserves gland topology, and \(\mathcal{L}_{\mathrm{Boundary}}\)~\cite{kervadec2019boundary} improves boundary accuracy.

\paragraph{Unsupervised loss.}
For unlabeled images, the student learns from the refined pseudo-mask \(y_r\) under confidence weighting \(w\):

\begin{equation}
\mathcal{L}_{\mathrm{unsup}}^{\mathrm{refine}}
=
\mathbb{E}\!\left[
w \cdot \left(\mathcal{L}_{\mathrm{BCE}}(\hat{y}_u^w, y_r) + \mathcal{L}_{\mathrm{Dice}}(\hat{y}_u^w, y_r)\right)
\right]
\end{equation}
To encourage augmentation-invariant representations, a consistency loss is imposed between weakly and strongly augmented student predictions:

\begin{equation}
\mathcal{L}_{\mathrm{unsup}}^{\mathrm{cons}}
=
\mathcal{L}_{\mathrm{BCE}}(\hat{y}_u^s, \hat{y}_u^w) + \mathcal{L}_{\mathrm{Dice}}(\hat{y}_u^s, \hat{y}_u^w)
\end{equation}
The full unsupervised objective is
\begin{equation}
\mathcal{L}_{\mathrm{unsup}}
=
\lambda_u \mathcal{L}_{\mathrm{unsup}}^{\mathrm{refine}}
+
\lambda_c \mathcal{L}_{\mathrm{unsup}}^{\mathrm{cons}}
\end{equation}

\section{Experiments}


\subsection{Datasets}
We evaluate the proposed framework on the GlaS~\cite{sirinukunwattana2017gland} and CRAG~\cite{graham2019mild} histopathology gland segmentation benchmarks, containing 165 and 213 images, respectively. For semi-supervised learning, 10\% and 20\% of the training data are randomly selected as labeled, with the remainder used as unlabeled. Main results use a fixed labeled split, while ablation studies report the mean $\pm$ std over three random seeds.

\subsection{Implementation Details}

All experiments are conducted on a single NVIDIA RTX PRO 4000 Blackwell GPU (24GB). Images are resized to \(256\times256\) for training, while inference is performed on non-overlapping patches and stitched to reconstruct the full-resolution mask. Models are trained for 100 epochs using AdamW (\(1\times10^{-4}\), batch size 16), with EMA teacher updates (\(\alpha=0.99\)). We set \(\lambda_1=\lambda_2=1.0\), \(\lambda_3=\lambda_4=0.5\), \(\lambda_u=\lambda_c=0.25\), and use a 10-epoch warmup.

For diffusion-based refinement, we use \(T = 100\) timesteps with a partial noising ratio of \(\alpha_{\mathrm{n}} = 0.2\). Low-confidence pixels are identified using a FreeMatch-style adaptive threshold, where \(c=\max(p,1-p)\). The threshold \(\tau\) is updated as an EMA of the batch mean of the maximum teacher confidence for each image and clamped to \([\tau_{\min}, \tau_{\max}] = [0.60, 0.75]\), where \(\tau_{\min}\) and \(\tau_{\max}\) prevent overly conservative and aggressive refinement, respectively. At inference, we use a fixed threshold of \(\tau=0.75\). During inference, only the student network is used, and performance is evaluated using mDice and mJaccard.

\subsection{Comparison with State of the Art}
\begin{table}[t!]
\centering
\caption{Comparison with state-of-the-art semi-supervised segmentation methods on the GlaS and CRAG datasets. The best performance is highlighted in \textbf{bold}, and the second-best among all compared methods is \underline{underlined}. Baseline results are taken from UniSemAlign~\cite{van2026unisemalign}, which uses the same experimental protocol. The last row reports our proposed method.}
\label{tab:ssl_sota}

\scriptsize
\setlength{\tabcolsep}{4pt}
\renewcommand{\arraystretch}{1.08}

\begin{tabular}{l|l|cc|cc}
\toprule

\makecell[c]{Labeled\\Ratio} &
Method &
\multicolumn{2}{c|}{GlaS} &
\multicolumn{2}{c}{CRAG} \\
\cmidrule(lr){3-4}
\cmidrule(lr){5-6}

&
&
mDice (\%) &
mJaccard (\%) &
mDice (\%) &
mJaccard (\%) \\
\midrule

100\% &
Fully-Supervised
& 89.59 & 81.84
& 92.68 & 86.36 \\
\midrule

\multirow{10}{*}{10\%}
& UAMT (MICCAI'19) \cite{yu2019uncertainty}     & 78.57 & 64.70 & 77.85 & 63.74 \\
& FixMatch (NeurIPS'20) \cite{sohn2020fixmatch} & 66.01 & 51.39 & 72.42 & 64.31 \\
& CPS (CVPR'21) \cite{chen2021semi}        & 61.70 & 49.34 & 47.10 & 40.77 \\
& CT (MIDL'22) \cite{luo2022semi}          & 81.75 & 70.44 & 77.33 & 65.56 \\
& XNet (ICCV'23)  \cite{xnetv2_2023}      & 74.13 & 58.90 & 65.27 & 50.40 \\
& CorrMatch (CVPR'24) \cite{sun2024corrmatch}  & 85.54 & 74.74
                         & 79.93 & 66.76 \\
& DuSSS (AAAI'25) \cite{pan2024dusss}      & 75.07 & 61.46 & 64.25 & 49.99 \\
& CSDS (MICCAI-W'25) \cite{pmlr-v316-pham26a}     & 82.89 & 71.61 & 79.86 & 67.81 \\
& UniSemAlign (CVPRW'26) \cite{van2026unisemalign}
                         & \textbf{88.15} & \underline{78.82}
                         & \underline{88.57} & \underline{79.52} \\
& \textbf{CoDiR (Ours)}
                         & \underline{88.09} & \textbf{79.41}
                         & \textbf{89.83} & \textbf{82.46} \\
\midrule

\multirow{10}{*}{20\%}
& UAMT (MICCAI'19) \cite{yu2019uncertainty}     & 77.09 & 62.72 & 79.76 & 66.33 \\
& FixMatch (NeurIPS'20) \cite{sohn2020fixmatch} & 58.51 & 44.12 & 71.20 & 64.75 \\
& CPS (CVPR'21) \cite{chen2021semi}        & 75.99 & 63.69 & 52.43 & 45.70 \\
& CT (MIDL'22) \cite{luo2022semi}         & 85.96 & 76.51
                         & 81.64 & 70.87 \\
& XNet (ICCV'23) \cite{xnetv2_2023}       & 79.56 & 67.50 & 65.60 & 50.85 \\
& CorrMatch (CVPR'24) \cite{sun2024corrmatch}   & 86.09 & 75.59 & 85.35 & 74.54 \\
& DuSSS (AAAI'25) \cite{pan2024dusss}      & 79.50 & 66.80 & 71.13 & 57.45 \\
& CSDS (MICCAI-W'25) \cite{pmlr-v316-pham26a}      & 83.50 & 72.87 & 81.28 & 69.67 \\
& UniSemAlign (CVPRW'26) \cite{van2026unisemalign}
                         & \underline{88.58} & \underline{79.50}
                         & \underline{89.40} & \underline{80.88} \\
& \textbf{CoDiR (Ours)}
                         & \textbf{89.19} & \textbf{81.09}
                         & \textbf{90.29} & \textbf{82.93} \\
\bottomrule

\end{tabular}
\end{table}
\noindent\textbf{Quantitative Results.} As shown in Table~\ref{tab:ssl_sota}, CoDiR achieves the best result on seven of the eight benchmark metrics. Relative to the strongest competing method, our earlier UniSemAlign framework~\cite{van2026unisemalign}, CoDiR improves by \(+1.26\%\) in mDice and \(+2.94\%\) in mJaccard on CRAG under 10\% labeling, and is on par on GlaS (\(-0.06\%\) in mDice, \(+0.59\%\) in mJaccard). Under 20\% labeling the gains are \(+0.61\%\) in mDice and \(+1.59\%\) in mJaccard on GlaS, and \(+0.89\%\) in mDice and \(+2.05\%\) in mJaccard on CRAG. The improvement is consistently larger on CRAG, whose glands are larger and less regular, so that pseudo-labels there contain more structural error for the refinement step to correct.\\

\begin{figure*}[!t]
    \centering
    \includegraphics[
        width=\textwidth,
        trim=0cm 0cm 0cm 0cm,
        clip
    ]{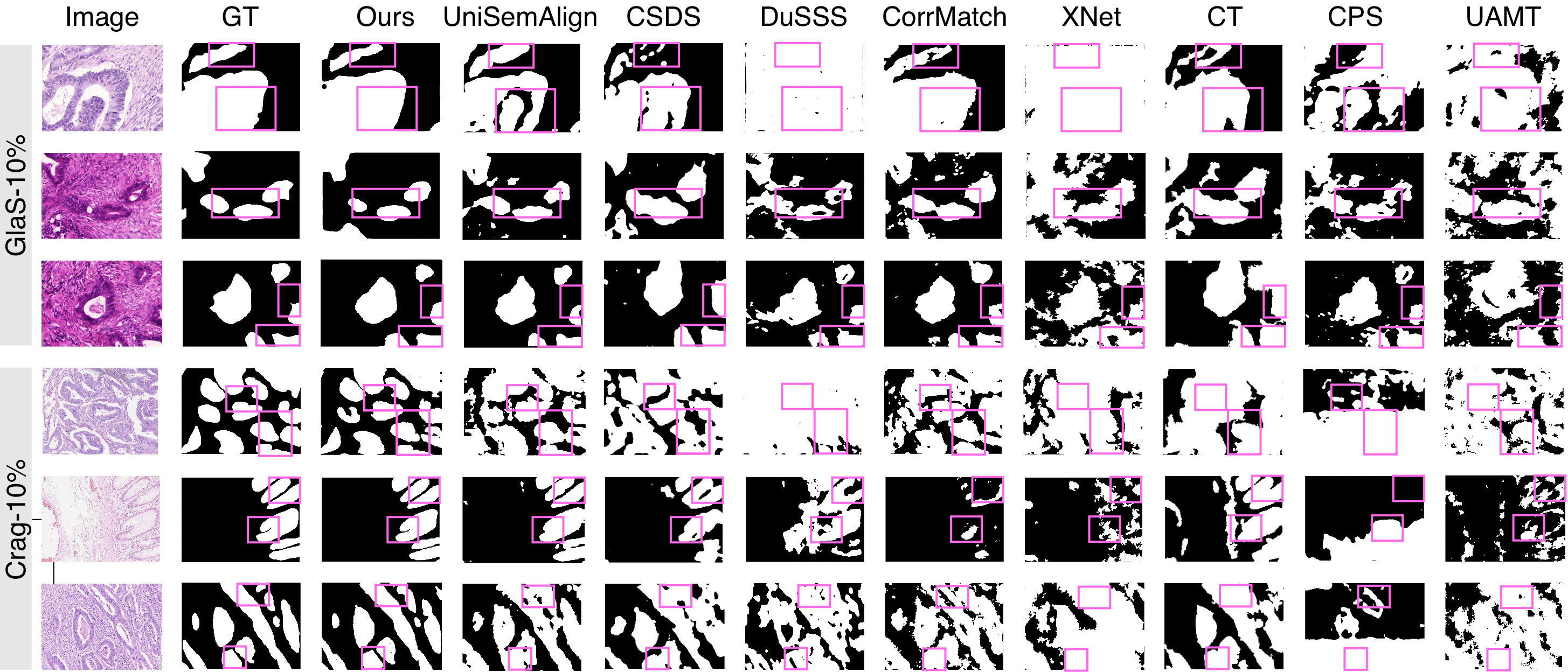}
    \caption{Predicted masks at the 10\% labeling ratio: GlaS in the upper three rows, CRAG in the lower three. Pink rectangles mark the regions where the compared methods diverge most visibly.}
    \label{fig:qualitative}
\end{figure*}

\noindent\textbf{Qualitative Results.}
Fig.~\ref{fig:qualitative} compares predictions under the 10\% labeling setting. CoDiR delineates gland boundaries with fewer spurious foreground regions than the baselines and retains narrow structures that competing methods break apart, yielding masks closer to the ground truth. Both effects are concentrated in the ambiguous areas that the refinement step targets.

\subsection{Ablation Studies}
\label{sec:ablation}
All ablation studies are conducted on the GlaS dataset under the 10\% labeling setting, with results reported as mean $\pm$ std over three random seeds.

\noindent\textbf{Effect of Visual Backbone.}
Table~\ref{tab:ablation_backbone} shows that UNI consistently outperforms the other visual encoders, achieving 88.09\% mDice and 79.41\% mJaccard. This demonstrates the advantage of using a pathology-specific visual representation for histopathology segmentation.

\begin{table}[H]
\centering
\caption{Ablation study of different visual encoders in the Mean Teacher framework.}
\label{tab:ablation_backbone}

\scriptsize
\setlength{\tabcolsep}{4pt}
\renewcommand{\arraystretch}{1.15}

\begin{tabular}{lcc}
\toprule
Backbone & mDice (\%) & mJaccard (\%) \\
\midrule
ResNet101 & $65.93\pm0.15$ & $51.10\pm0.13$ \\
ResNet200 & $74.75\pm0.10$ & $60.64\pm0.11$ \\
MedCLIP   & $72.65\pm0.17$ & $58.47\pm0.23$ \\
CONCH     & $78.89\pm0.04$ & $66.55\pm0.14$ \\
UNI       & $\mathbf{88.09\pm0.32}$ & $\mathbf{79.41\pm0.46}$ \\
\bottomrule
\end{tabular}
\end{table}

\noindent\textbf{Effect of Proposed Modules.}
Table~\ref{tab:ablation_modules} shows the incremental contribution of each CoDiR module. Diffusion Refinement provides the largest gain, while Confidence Fusion and Confidence Weighting further improve the performance, resulting in 88.09\% mDice and 79.41\% mJaccard.

\begin{table}[H]
\centering
\caption{Incremental ablation study of the proposed CoDiR modules, starting from a Mean Teacher baseline with the UNI encoder and DeepLabV3 decoder.}
\label{tab:ablation_modules}

\scriptsize
\setlength{\tabcolsep}{4pt}
\renewcommand{\arraystretch}{1.15}

\begin{tabular}{lcc}
\toprule
Configuration & mDice (\%) & mJaccard (\%) \\
\midrule
Mean Teacher (UNI) &
$80.95\pm0.12$ & $69.09\pm0.79$ \\

+ Diffusion Refinement &
$87.31\pm0.26$ & $78.28\pm0.40$ \\

+ Confidence Fusion &
$87.50\pm0.31$ & $78.54\pm0.47$ \\

+ Confidence Weighting &
$\mathbf{88.09\pm0.32}$ & $\mathbf{79.41\pm0.46}$ \\
\bottomrule
\end{tabular}
\end{table}

\noindent\textbf{Effect of Confidence Threshold.}
As shown in Table~\ref{tab:ablation_tau}, varying the confidence threshold ceiling $\tau_{\text{max}}$ (with $\tau_{\min}=0.60$ fixed) yields stable performance, with $\tau_{\text{max}}=0.75$ performing best; this value is adopted for both training and inference. These results indicate that CoDiR remains stable across different choices of $\tau_{\text{max}}$.

\begin{table}[!h]
\centering
\caption{Effect of the confidence threshold $\tau_{\text{max}}$}
\label{tab:ablation_tau}

\scriptsize
\setlength{\tabcolsep}{10pt}
\renewcommand{\arraystretch}{1.15}

\begin{tabular}{lcc}
\toprule
$\tau_{\text{max}}$ & mDice (\%) & mJaccard (\%) \\
\midrule
$0.70$ & $87.70 \pm 0.05$ & $78.89 \pm 0.16$ \\
$0.75$ & \textbf{88.09 ±  0.32} & \textbf{79.41 ±  0.46} \\
$0.80$ & $87.98 \pm 0.39$ & $79.29 \pm 0.62$ \\
\bottomrule
\end{tabular}
\end{table}
\noindent\textbf{Effect of Noise Level.}
Table~\ref{tab:noise_level} analyzes the impact of the noise level $\alpha_{\mathrm{n}}$ in diffusion refinement. Moderate noise ($\alpha_{\mathrm{n}}=0.20$) achieves the best performance, while both smaller and larger values lead to slight degradation. This indicates that an appropriate noise level is important for effective pseudo-label refinement.

\begin{table}[H]
\centering
\caption{Sensitivity analysis of the diffusion noise level $\alpha_{\mathrm{n}}$.}
\label{tab:noise_level}
\scriptsize
\setlength{\tabcolsep}{4pt}
\renewcommand{\arraystretch}{1.1}
\begin{tabular}{ccc}
\toprule
$\alpha_{\mathrm{n}}$ & mDice (\%) & mJaccard (\%) \\
\midrule
0.05 & $87.07\pm0.09$ & $77.88\pm0.14$ \\
0.10 & $87.98\pm0.39$ & $79.29\pm0.62$ \\
0.20 & $\mathbf{88.09\pm0.32}$ & $\mathbf{79.41\pm0.46}$ \\
0.40 & $87.61\pm0.69$ & $78.67\pm0.05$ \\
0.60 & $87.86\pm0.38$ & $79.05\pm0.33$ \\
\bottomrule
\end{tabular}
\end{table}

\noindent\textbf{Effect of EMA Decay.}
As shown in Table~\ref{tab:ema_decay}, the performance remains relatively stable across different EMA decay rates. Among the evaluated settings, \(d=0.99\) achieves the best performance and is therefore adopted in all experiments.

\begin{table}[H]
\centering
\caption{Sensitivity analysis of the EMA decay rate $d$.}
\label{tab:ema_decay}
\scriptsize
\setlength{\tabcolsep}{4pt}
\renewcommand{\arraystretch}{1.1}
\begin{tabular}{ccc}
\toprule
$d$ & mDice (\%) & mJaccard (\%) \\
\midrule
0.90 & $87.65\pm0.18$ & $78.78\pm0.19$ \\
0.99 & $\mathbf{88.09\pm0.32}$ & $\mathbf{79.41\pm0.46}$ \\
0.995 & $87.22\pm0.05$ & $78.14\pm0.13$ \\
0.999 & $87.42\pm0.06$ & $78.35\pm0.13$ \\
\bottomrule
\end{tabular}
\end{table}






\section{Conclusion}
In this work, we presented CoDiR, a semi-supervised histopathology segmentation framework that combines Mean Teacher learning with diffusion-based pseudo-label refinement. By selectively refining low-confidence regions while preserving reliable teacher predictions, CoDiR improves segmentation performance under limited annotation. Experiments on GlaS and CRAG demonstrate consistent improvements over recent semi-supervised methods, highlighting the effectiveness of the proposed framework. However, the diffusion refiner is trained on a relatively small labeled subset and evaluated only on two colorectal gland datasets, leaving its cross-organ and cross-stain generalization underexplored. Future work will focus on developing more data-efficient and structure-aware refinement strategies to improve boundary precision and enhance uncertain-region correction across diverse domains.

%
%
%
\bibliographystyle{splncs04}
\bibliography{refs/mybib}

\end{document}